Accepted by ISPRS Journal of Photogrammetry and Remote Sensing

# A Survey on Object Detection in Optical Remote Sensing Images

Gong Cheng, Junwei Han*

*School of Automation, Northwestern Polytechnical University, Xi'an, 710072, China*

*Corresponding author. Email: Junweihan2010@gmail.com*

**Abstract**: Object detection in optical remote sensing images, being a fundamental but challenging problem in the field of aerial and satellite image analysis, plays an important role for a wide range of applications and is receiving significant attention in recent years. While enormous methods exist, a deep review of the literature concerning generic object detection is still lacking. This paper aims to provide a review of the recent progress in this field. Different from several previously published surveys that focus on a specific object class such as building and road, we concentrate on more generic object categories including, but are not limited to, road, building, tree, vehicle, ship, airport, urban-area. Covering about 270 publications we survey 1) template matching-based object detection methods, 2) knowledge-based object detection methods, 3) object-based image analysis (OBIA)-based object detection methods, 4) machine learning-based object detection methods, and 5) five publicly available datasets and three standard evaluation metrics. We also discuss the challenges of current studies and propose two promising research directions, namely deep learning-based feature representation and weakly supervised learning-based geospatial object detection. It is our hope that this survey will be beneficial for the researchers to have better understanding of this research field.

## 1. Introduction

Object detection in optical remote sensing images (RSIs) is to determine if a given aerial or satellite image contains one or more objects belonging to the class of interest and locate the position of each predicted object in the image. The term 'object' used in this survey refers to its generalized form, including man-made objects (e.g. vehicles, ships, buildings, etc.) that have sharp boundaries and are independent of background environment, as well as landscape objects, such as land-use/land-cover (LULC) parcels that have vague boundaries and are parts of background environment. As a fundamental problem in the field of aerial and satellite image analysis, object detection in optical RSIs plays an important role for a wide range of applications, such as environmental monitoring, geological hazard detection, LULC mapping, geographic information system (GIS) update, precision agriculture, urban planning, etc.

Object detection in optical RSIs often suffers from several increasing challenges including the large variations in the visual appearance of objects caused by viewpoint variation, occlusion, background clutter, illumination, shadow, etc., the explosive growth of RSIs in quantity and quality, and the various requirements of new application areas. To address these challenges, the topic of geospatial object detection has been extensively studied since the 1980s. The low spatial resolution of earlier satellite images (such as Landsat) would not allow the detection of separate man-made or natural objects. Therefore, researchers mostly focused on extracting the region properties from these images. With the advances of remote sensing technology, the very high resolution (VHR) satellite (e.g. IKONOS, SPOT-5, and Quickbird) and aerial images have been providing us more detailed spatial and textural information. Aside from region properties, a greater range of man-made objects become recognizable and even can be separately identified than ever before because of the increased sub-meter resolution. This opens new prospects in the field of automatic detection of geospatial objects.

During the last decades, considerable efforts have been made to develop various methods for the detection of different types of objects in satellite and aerial images, such as roads (Barsi and Heipke, 2003; Barzohar and Coope, 1996; Chaudhuri et al., 2012; Das et al., 2011; Hu et al., 2007; Huang and Zhang, 2009; Kim et al., 2004; Laptev et al., 2000; Leninisha and Vani, 2015; Li et al., 2010; Maillard and Cavayas, 1989; Mayer et al., 2006; McKeown and Denlinger, 1988; Mena, 2003; Mokhtarzade and Zoej, 2007; Movaghati et al., 2010; Song and Civco, 2004; Trinder and Wang, 1998; Ünsalan and Sirmacek, 2012; Wang et al., 2015; Wang and Zhang, 2011; Zhang et al., 2011a; Zhang and Couloigner, 2006; Zhou et al., 2006; Zhu et al., 2005), buildings (Ahmadi et al., 2010; Akçay and Aksoy, 2010; Aytekın et al., 2012; Benedek et al., 2012; Durieux et al., 2008; Hofmann et al., 2002; Karantzalos and Paragios, 2009; Lefèvre et al., 2007; Lhomme et al., 2009; Mayer, 1999; Ok, 2013; Ok et al., 2013; Peng and Liu, 2005; Peng et al., 2005; Senaras et al., 2013; Shufelt, 1999; Sirmaçek and Ünsalan, 2009; Sirmacek and Ünsalan, 2011; Stankov and He, 2013, 2014; Wegner et al., 2011a; Wegner et al., 2011b), trees (Haala and Brenner, 1999; Hung et al., 2012; Malek et al., 2014; Moustakidis et al., 2012; Yang et al., 2013), vehicles (Eikvil et al., 2009; Grabner et al., 2008; Jin and Davis, 2007; Kembhavi et al., 2011; Leitloff et al., 2010; Moon et al., 2002; Niu, 2006; Tuermer et al., 2013; Wen et al., 2015; Yang et al., 2015; Yang et al., 2013; Zhao and Nevatia, 2003; Zheng et al., 2013), etc. While enormous methods exist, a deep review of the literature concerning generic object detection is still lacking. This paper aims to provide a review of the recent progress in this field. Different from several previously published surveys that focus on a specific object class such as building (Mayer, 1999; Shufelt, 1999) and road (Mayer et al., 2006; Mena, 2003), we concentrate on more generic object categories including, but are not limited to, road, building, tree, vehicle, ship, airport, urban-area. Covering about 270 publications we survey 1) template matching-based object detection methods, 2) knowledge-based object detection methods, 3) object-based image analysis (OBIA)-based object detection methods, 4) machine learning-based object detection methods, and 5) five publicly available datasets and three standard evaluation metrics for object detection. We also discuss open problems and challenges of current studies, and propose two promising research directions in future for constructing more effective object detection framework. This survey will be especially beneficial for the researchers to have better understanding of this research field. Furthermore, to the best of our knowledge, this is the first survey paper in the literature that focuses on generic object detection in optical RSIs.

The rest of the paper is organized as follows. Section 2 briefly introduces the taxonomy of methods for object detection. In Sections 3, 4, 5 and 6, we exhaustively review template matching-based object detection methods, knowledge-based object detection methods, OBIA-based object detection methods, and machine learning-based object detection methods, respectively. In Section 7, we review five publicly available datasets and three standard evaluation metrics. Section 8 discusses the challenges of current studies and proposes two promising research directions to advance the field. Finally, conclusions are drawn in Section 9.

## 2. Taxonomy of methods for object detection

In the last decades, a large number of methods have been developed for object detection from aerial and satellite images. We can generally divide them into four main categories: template matching-based methods, knowledge-based methods, OBIA-based methods, and machine learning-based methods. These four categories are not necessarily independent and sometimes the same method exists with different categories. Fig. 1 shows a taxonomy of geospatial object detection studies, in which rounded rectangles with solid borders illustrate our scope in this paper.

According to the template type selected by a user, template matching-based methods are further divided into two classes as rigid template matching and deformable template matching. As for knowledge-based object detection methods, we mainly review two kinds of most widely used prior knowledge, namely geometric information and context information. Generally, OBIA-based object detection methods involve two steps: image segmentation and object classification. With regards to machine learning-based methods, we mainly focus on reviewing three crucial steps that play important roles in the performance of object detection. They are feature extraction, optional feature fusion and dimension reduction, and classifier training. In the step of feature extraction, we introduce five types of typical features including Histogram of oriented gradients (HOG) feature, bag-of-words (BoW) feature, texture features, sparse representation (SR)-based features, and Haar-like features. In the step of classifier training, we exhaustively review six kinds of machine learning algorithms including support vector machine (SVM), AdaBoost, $k$ -nearest-neighbor ($k$NN), conditional random field (CRF), sparse representation-based classification (SRC), and artificial neural network (ANN).

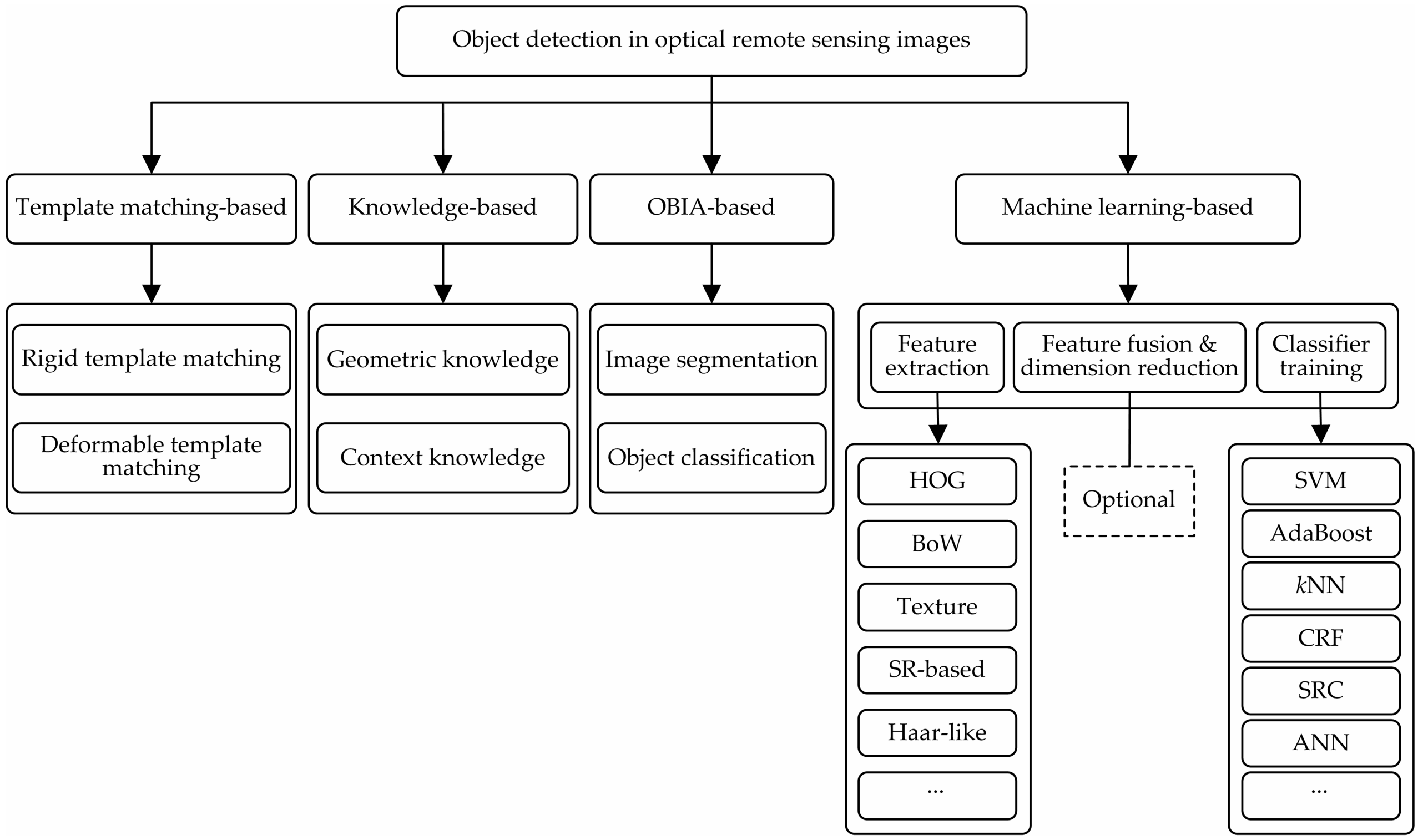

Fig. 1 Taxonomy of methods for object detection in optical RSIs. Rounded rectangles with solid borders illustrate our scope in this paper.

## 3. Template matching-based object detection

Template matching-based methods are one category of the simplest and earliest approaches for object detection. Fig. 2 gives the flowchart of template matching-based object detection. As shown in Fig. 2, there are two main steps in template matching-based object detection framework. 1) Template generation: a template $T$ for each to-be-detected object class should be firstly generated by hand-crafting or learning from the training set. 2) Similarity measure: given a source image, the stored template $T$ is used to match the image at each possible position to find the best matches, according to the minimum distortion or maximum correlation measures, while taking into account all allowable translation, rotation, and scale changes. The most popular similarity measures are the sum of absolute differences (SAD), the sum of squared differences (SSD), the normalized cross correlation (NCC), and the Euclidean distance (ED).

According to the template type selected by a user, the template matching-based object detection approaches are generally categorized into two groups: rigid template matching and deformable template matching.

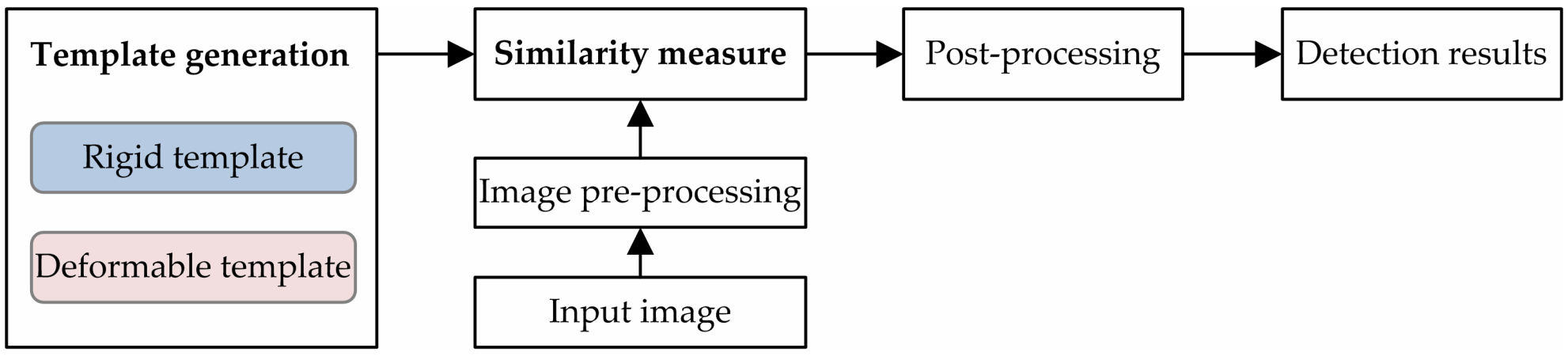

Fig. 2 The flowchart of template matching-based object detection.

### 3.1 Rigid template matching

Early research in this area mainly concentrated on rigid template matching. Various rigid templates have been designed for detecting specific objects with simple appearance and small variations such as road (Chaudhuri et al., 2012; Kim et al., 2004; McKeown and Denlinger, 1988; Zhang et al., 2011a; Zhou et al., 2006). For instance, McKeown and Denlinger (1988) introduced a road tracking method based on road profile correlation, in which the distortion of a reference profile and the target profile is measured by identifying two geometric parameters (shift and width) and two radiometric parameters (brightness and contrast). Zhou et al. (2006) introduced a road tracking system by using two profiles: one orthogonal to the road direction and the other parallel to the road direction. Kim et al. (2004) utilized a rectangular template instead of the profiles to track ribbon roads in urban areas via least squares correlation template matching method. Zhang et al. (2011a) proposed a semi-automatic template matching method to track roads, in which a rectangular reference template is generated by user inputting a seed point on a selected road and using a spoke wheel algorithm to obtain road direction, road width, and starting point. The matching is then performed by calculating the SSD and Euclidean distance transformation of the reference template and a target template.

The morphological hit-or-miss transform (HMT) is a powerful tool dedicated to template matching. Since its first definition for binary images (Lefèvre et al., 2007), its recent extensions to grayscale images (Stankov and He, 2013, 2014; Zheng et al., 2013) and multispectral images (Weber and Lefèvre, 2008, 2012) have proven its ability to solve various template matching problems. Lefèvre et al. (2007) presented a method for building extraction in Quickbird images based on an adaptive binary HMT with varying sizes and shapes of the structuring element. In this method, a binary image was firstly generated from a gray-level panchromatic input image before building detection. To make use of the spectral information, the authors in (Stankov and He, 2013, 2014) generated grayscale images from the spectral bands and then applied the grayscale HMT to building detection, where (Stankov and He, 2013) is a supervised method that requires a set of reference windows for each roof color present in the image while the method in (Stankov and He, 2014) is unsupervised. Furthermore, Weber and Lefèvre (2008, 2012) introduced a new definition of the HMT for multivariate image analysis and illustrated its potential as a template matching operator for coastline extraction and petroleum tank detection.

However, although the rigid template matching is effective in some applications, it has a number of disadvantages, resulting in their utility is limited. A shortcoming of the rigid template matching is that it requires the template to be very precise, so it is sensitive to shape and density variation. In most applications, an exact geometric template of the object is not available because of the viewpoint change, or large intra-class variations among the objects. For instance, most of the road trackers mentioned above do not work well when they encounter irregular geometric deformation in tracking process, such as appearances of road junctions, material changes, occlusions of cars, shadows and lane markings, etc.

## 3.2 Deformable template matching

The concept of deformable template was first introduced to computer vision community by Fischler and Elschlager (1973) with the spring-loaded templates. Deformable template matching is more powerful and flexible than rigid shape matching in dealing with shape deformations and intra-class variations because of its capability to both impose geometrical constraints on the shape and to integrate local image evidence. There has been a substantial amount of studies on deformable template matching in recent years. These studies can be roughly divided into two classes: free-form deformable templates and parametric deformable templates (Jain et al., 1998).

### *3.2.1 Free-form deformable templates*

The free-form deformable templates represent an arbitrary object shape by constraining some general regularization (e.g. continuity, smoothness, etc.) and the most popular methods are the active contour models, also known as snake models. In these approaches, an energy-minimizing contour, called a "snake", is controlled by a combination of three forces or energies. Briefly, a snake is modeled as being able to deform elastically, but any deformation increases its internal energy causing a "restitution force", which tries to bring it back to its original shape (Jain et al., 1998). For different applications, various snake-based methods have been developed (Ahmadi et al., 2010; Jing et al., 2011; Laptev et al., 2000; Leninisha and Vani, 2015; Liu et al., 2013b; Niu, 2006; Peng et al., 2005; Wang and Zhang, 2011; Xu and Duan, 2010). Peng et al. (2005) proposed an improved snake model for building detection from gray-level aerial images by modifying the traditional snake model in two aspects: the criteria for the selection of initial seeds and the external energy function. Niu (2006) introduced a semi-automatic framework based on geometric active contour model for highway extraction and vehicle detection from aerial photographs. In (Liu et al., 2013b), the authors proposed an algorithm to extract geospatial objects with regular shape using shape-based global minimization active contour model. Xu and Duan (2010) described a shape-matching approach for aircraft recognition at low altitude, in which an artificial bee colony (ABC) algorithm with edge potential function (EPF) was proposed.

### *3.2.2 Parametric deformable templates*

The parametric deformable templates parameterize a specific object class and its variations by a parametric formula (Barzohar and Coope, 1996; Hung et al., 2012; Lhomme et al., 2009; Liu et al., 2013a; Movaghati et al., 2010) or using a prototype and deformation modes (Karantzalos and Paragios, 2009; Lin et al., 2015; Sirmaçek and Ünsalan, 2009; Sun et al., 2010; Tao et al., 2011), which are commonly used when some prior information of the geometrical shape is available. For instance, Lhomme et al. (2009) proposed a “Discrimination by Ratio of Variance” (DRV) parameter to quantify the spatial distribution of grey-level variation of a building and its close neighborhood for the detection of buildings. Liu et al. (2013a) proposed an aircraft recognition method in high-resolution satellite images using coarse-to-fine shape prior, in which a parametric shape model for aircraft is derived by applying principal component analysis (PCA) and kernel density function. In (Sirmaçek and Ünsalan, 2009), the authors firstly defined two template building images, one for a bright building and the other one for a dark building, and represented them by scale invariant feature transform (SIFT) features (Lowe, 2004). Then, the urban area was extracted using a multiple subgraph matching method and separate buildings in the urban area were extracted using a graph cut method.

A brief summary of template matching-based object detection methods is given in Table 1.

Table 1 A brief summary of template matching-based object detection methods

| Templates | Objects and publications | Strengths | Limitations |
|---|---|---|---|
| Rigid template | building (Lefèvre et al., 2007; Stankov and He, 2013, 2014); coastline and storage tank (Weber and Lefèvre, 2008, 2012); road (Chaudhuri et al., 2012; Kim et al., 2004; McKeown and Denlinger, 1988; Zhang et al., 2011a; Zhou et al., 2006). | Simple and easy to implement | Scale and rotation dependent; Sensitive to shape and viewpoint change |
| Deformable template | airplane, ship, and storage tank (Lin et al., 2015; Liu et al., 2013a, b; Sun et al., 2010; Xu and Duan, 2010); airport (Tao et al., 2011); building (Ahmadi et al., 2010; Karantzalos and Paragios, 2009; Sirmaçek and Ünsalan, 2009); edges of oil slick (Jing et al., 2011); road (Barzohar and Coope, 1996; Laptev et al., 2000; Leninisha and Vani, 2015; Lhomme et al., 2009; Movaghati et al., 2010; Niu, 2006; Peng et al., 2005; Wang and Zhang, 2011); tree crown (Hung et al., 2012). | More powerful and flexible than rigid shape matching in dealing with shape deformations and intra-class variations | Need more prior information and parameters of the geometrical shape for template designing; Computationally expensive |

## 4. Knowledge-based object detection

Knowledge-based object detection methods are another type of popular approaches for object detection in optical RSIs. An extensive collection of papers on knowledge-based object detection have been published for buildings in (Akçay and Aksoy, 2010; Haala and Brenner, 1999; Hofmann et al., 2002; Huertas and Nevatia, 1988; Irvin and McKeown, 1989; Lin and Nevatia, 1998; Liow and Pavlidis, 1990; McGlone and Shufelt, 1994; Ok, 2013; Ok et al., 2013; Peng and Liu, 2005; Shufelt, 1996; Stilla et al., 1997; Weidner and Förstner, 1995), roads in (Hu et al., 2007; Leninisha and Vani, 2015; Maillard and Cavayas, 1989; Trinder and Wang, 1998; Wang and Newkirk, 1988; Zhu et al., 2005), and other more general object extraction applications like landslide, bridges, vehicles, urban land changes, crops, drainage channels, forests in (Chaudhuri and Samal, 2008; Janssen and Middelkoop, 1992; Martha et al., 2011; Moon et al., 2002; Solberg, 1999; Tchoku et al., 1996; Wang, 1993). A review of knowledge-based object extraction in RSIs is also given in (Baltsavias, 2004). This type of approaches generally translates object detection problem into hypotheses testing problem by establishing various knowledge and rules. Fig. 3 gives the flowchart of knowledge-based object detection. As shown in Fig. 3, the establishment of knowledge and rules is the most important step. Two kinds of widely used knowledge on target objects are geometric knowledge (Huertas and Nevatia, 1988; McGlone and Shufelt, 1994; Shufelt, 1996; Trinder and Wang, 1998; Weidner and Förstner, 1995) and context knowledge (Akçay and Aksoy, 2010; Huertas and Nevatia, 1988; Irvin and McKeown, 1989; Lin and Nevatia, 1998; Ok, 2013; Ok et al., 2013; Peng and Liu, 2005; Stilla et al., 1997).

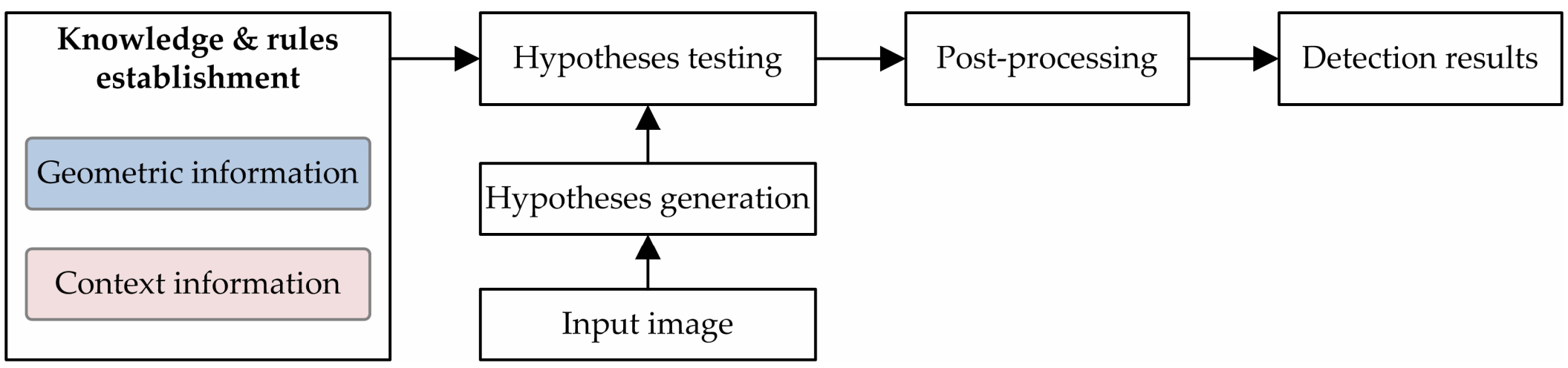

Fig. 3 The flowchart of knowledge-based object detection.

### 4.1 Geometric knowledge

The object geometric information is the most important and widely used knowledge for object detection, which encodes prior knowledge by taking parametric specific or generic shape models (Huertas and Nevatia, 1988; Leninisha and Vani, 2015; McGlone and Shufelt, 1994; Trinder and Wang, 1998; Weidner and Förstner, 1995). For example, Trinder and Wang (1998) proposed a road model including geometric and radiometric properties, in which the hypotheses of roads are generated using hand-crafted rules, and a top-down process is applied to verify the road hypotheses. Huertas and Nevatia (1988) assumed that the buildings are rectangular or composed of rectangular components (e.g. "box," "T," " L," and "E" shapes) and used a generic model of the shapes to detect buildings. Weidner and Förstner (1995) developed an approach for extracting 3D shape of buildings from high-resolution digital elevation models (DEMs) by establishing and using explicit geometric constraints knowledge in the form of parametric and prismatic building models. McGlone and Shufelt (1994) proposed to include the target geometric and metric knowledge into the building extraction system for the generation of building hypotheses, and the generated hypotheses were finally verified with the shadow information.

### 4.2 Context knowledge

The context knowledge is another crucial cue for knowledge-based object detection and the most widely used context knowledge is the spatial constraints or relationships between objects and background, or the information regarding how the object interacts with its neighboring regions (Akçay and Aksoy, 2010; Chaudhuri and Samal, 2008; Liow and Pavlidis, 1990; Ok et al., 2013; Peng and Liu, 2005). As a representative example, shadow evidence has been considered to be one of the most important clues for building detection (Irvin and McKeown, 1989; Lin and Nevatia, 1998; Liow and Pavlidis, 1990; Ok, 2013; Ok et al., 2013). For instance, Irvin and McKeown (1989) exploited the relationship between man-made structures and their cast shadows to predict the locations and shape of buildings, and Liow and Pavlidis (1990) used shadow information to complete the boundary grouping process. Ok et al. (2013) employed the shadow evidence to automatically detect buildings with arbitrary shapes from monocular VHR images. In their work, the authors modeled the directional spatial relationship between buildings and their shadows by proposing a new fuzzy landscape generation approach. Besides, the shadow areas derived from monocular images were extensively used during the verification of the building hypotheses. For example, Lin and Nevatia (1998) proposed an approach to detect buildings from oblique aerial images, in which the hypothesized rectangular buildings are verified with both shadow and wall evidences by following the basic assumption that the building shapes are rectilinear. Peng and Liu (2005) developed a shadow-context model to extract buildings in dense urban aerial images via combining shadow information with context to verify building regions.

It is worth noting that the core of knowledge-based object detection methods is how to effectively transform the implicit knowledge understanding on target objects into the explicit detection rules. If the defined rules are too strict, some target objects will be missed; conversely, too loose rules will cause false positives. A brief summary of knowledge-based object detection methods is given in Table 2.

Table 2 A brief summary of knowledge-based object detection methods

| Objects and publications | Strengths | Limitations |
|---|---|---|
| bridge (Chaudhuri and Samal, 2008); building (Akçay and Aksoy, 2010; Haala and Brenner, 1999; Hofmann et al., 2002; Huertas and Nevatia, 1988; Irvin and McKeown, 1989; Lin and Nevatia, 1998; Liow and Pavlidis, 1990; McGlone and Shufelt, 1994; Ok, 2013; Ok et al., 2013; Peng and Liu, 2005; Shufelt, 1996; Stilla et al., 1997; Weidner and Förstner, 1995); channel (Tchoku et al., 1996); crops (Janssen and Middelkoop, 1992); drainage forest (Solberg, 1999); landslide (Martha et al., 2011); road (Hu et al., 2007; Leninisha and Vani, 2015; Maillard and Cavayas, 1989; Trinder and Wang, 1998; Wang and Newkirk, 1988; Zhu et al., 2005); urban land change (Wang, 1993); vehicle (Moon et al., 2002). | Detection can be performed through coarse-to-fine hierarchical structure. | How to define prior knowledge and detection rules is subjective; Too loose rules will cause false positives and vice versa. |

## 5. OBIA-based object detection

Recently, with the increasing availability and wide utilization of sub-meter imagery, object-based image analysis (OBIA or GEOBIA for geospatial object based image analysis) (Blaschke, 2010; Blaschke et al., 2008) has become a new methodology or paradigm (Blaschke et al., 2014) to classify or map VHR imagery into meaningful objects (or rather, grouping of relatively local homogeneous pixels). Fig. 4 gives the flowchart of OBIA-based object detection. As shown in Fig. 4, OBIA involves two steps: image segmentation and object classification. Firstly, imagery is first segmented into homogeneous regions (segments also called objects) representing a relatively homogeneous group of pixels by selecting desired scale, shape, and compactness criteria. And in a second step, a classification process is applied to these objects. Since OBIA offers the potential to exploit geographical information system (GIS) functionality, such as the incorporation of the spatial context or object shape in the classification, it provides a framework for overcoming the limitations of conventional pixel-based image classification methods and has been successfully applied to landslide mapping (Feizizadeh et al., 2014; Li et al., 2015b; Martha et al., 2010; Martha et al., 2011; Martha et al., 2012; Stumpf and Kerle, 2011), land cover and land use mapping (Baker et al., 2013; Benz et al., 2004; Blaschke, 2003; Blaschke et al., 2011; Blaschke et al., 2008; Contreras et al., 2015; D'Oleire-Oltmanns et al., 2014; De Pinho et al., 2012; Doleire-Oltmanns et al., 2013; Drăguţ and Blaschke, 2006; Drăguţ and Eisank, 2012; Duro et al., 2012; Eisank et al., 2011; Goodin et al., 2015; Hay et al., 2003; Hofmann et al., 2011; Kim et al., 2011; Leon and Woodroffe, 2011; Li et al., 2014; Li and Shao, 2013; Lisita et al., 2013; Macfaden et al., 2012; Mallinis et al., 2008; Mishra and Crews, 2014; Moskal et al., 2011; Myint et al., 2011; Phinn et al., 2012; Tzotsos et al., 2011; Walker and Blaschke, 2008; Walker and Briggs, 2007; Weng, 2009, 2011; Xie et al., 2008; Yu et al., 2006; Zhou, 2013; Zhou et al., 2009; Zhou and Troy, 2008), and change detection (Bontemps et al., 2008; Chen and Hay, 2012; Contreras et al., 2016; Dissanska et al., 2009; Doxani et al., 2012; Doxani et al., 2008; Hussain et al., 2013; Im et al., 2008; Nebiker et al., 2014; Walter, 2004). Two reviews of the literature on OBIA/GEOBIA for remote sensing are also given in (Blaschke, 2010; Blaschke et al., 2014).

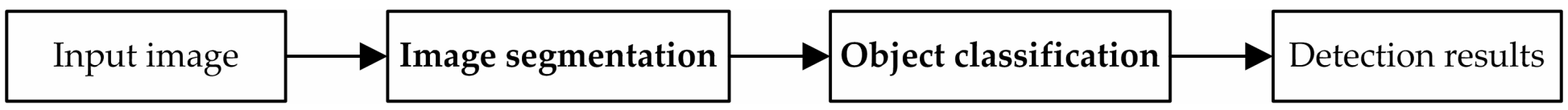

Fig. 4 The flowchart of OBIA-based object detection.

Image segmentation is the first step and a necessary prerequisite for generating the building blocks of OBIA based on a type of segmentation algorithm, the color, texture and shape of groups of pixels, and the required level or scale of spatial detail. The delineation quality of the target objects has a direct influence on the accuracy of the subsequent image classification. During the last decades, numerous image segmentation techniques have been developed and applied in RSI analysis (Ardila et al., 2012; Baatz and Schäpe, 2000; Benz et al., 2004; Blaschke et al., 2004; Drăguţ et al., 2014; Drăguţ et al., 2010; Esch et al., 2008; Gao et al., 2011; Hay et al., 2005; Jungho et al., 2014; Kim et al., 2008; Kim et al., 2011; Ming et al., 2015). However, as Hay et al. (2005) pointed out, the real challenge is to define appropriate segmentation parameters (typically based on spectral homogeneity, size, or both) for the varying sized, shaped, and spatially distributed image-objects composing a scene, so that segments can be generated to satisfy user requirements.

Multi-resolution segmentation (MRS) algorithm (Baatz and Schäpe, 2000) is probably the most popular one for the purpose of the delineation of relatively homogeneous and meaningful objects. Implemented in the eCognition® software, this algorithm has quickly became one of the most important segmentation algorithms within the OBIA domain. MRS uses three parameters to partition an image into objects: shape, compactness, and scale. The shape parameter defines to which percentage the homogeneity of shape is weighted against the homogeneity of spectral values. The compactness parameter is a sub-parameter of shape and is used to optimize image objects with regard to compactness or smoothness. The scale parameter is used for controlling the internal heterogeneity of the resulting objects and is therefore correlated with their average size, i.e., a larger value of the scale allows a higher internal heterogeneity, which increases the number of pixels per object and vice versa (Baatz and Schäpe, 2000; Benz et al., 2004; Flanders et al., 2003).

Since the scale is a crucial control parameter in MRS and heavily impacts on the classification accuracy,

how to select a more objective value of scale (at least traceable or reproducible) is a hot topic in OBIA (Blaschke, 2010; Blaschke et al., 2014). The traditional procedure for the selection of segmentation scale parameters often depends on subjective trial-and-error methods, which are mainly based on a visual assessment of segmentation suitability. While allowing flexibility in incorporating expert knowledge in OBIA, this procedure is hardly reproducible and raises some limitations with respect to the robustness of the approach. Based on the previous work of Woodcock and Strahler (1987), the concept of local variance (LV) graphs that reveals the spatial structure of images was introduced in the context of OBIA by Kim et al. (2008) to determine the optimal segmentation scale for alliance-level forest classification of multispectral IKONOS images. Drăguţ et al. (2010) carried the concept of LV further and presented a tool called Estimation of Scale Parameters (ESP) to objectively identify the most suitable range of scale parameters. In parallel, aiming at the automation of the segmentation process in OBIA, a suite of techniques (Ardila et al., 2012; Drăguţ et al., 2014; Drăguţ and Eisank, 2012; Esch et al., 2008; Hay et al., 2005) have also been developed.

Once segments are generated, one can extract object features, such as spectral information as well as size, shape, texture, geometry, and contextual semantic features. In addition, GIS-like functionality and expert knowledge can also be flexibly incorporated in classification procedures to make OBIA context-aware and multi-source capable (Blaschke et al., 2014). These features are then selected (optional, e.g. the work of Stumpf and Kerle (2011)) and fed to a classifier (e.g. membership function classifier, nearest neighbor classifier, decision tree, neural network, SVM, etc.) for classification.

For OBIA, accuracy assessment that measures how well objects extracted from high-resolution images match existing geographic objects is very important (Blaschke et al., 2014; Blaschke et al., 2008; Congalton and Green, 2009; Drăguţ et al., 2014; Drăguţ et al., 2010). It has been suggested by Congalton and Green (2009) that objects, instead of pixels, should be used as sampling units for the thematic accuracy assessment of object-based classification results. Thus, object-based accuracy assessment should account for the thematic accuracy of the class labels as well as the spatial characteristics of represented objects. Recently, a number of researchers have proposed several object-based metrics for assessing both geometric and thematic accuracies of object-based image classification (Albrecht, 2010; Clinton et al., 2010; Lizarazo, 2014; MacLean and Congalton, 2012; Radoux et al., 2011; Zhan et al., 2005; Zhen et al., 2013). However, accuracy assessments of OBIA are still a complex subject and some problems remain to be solved, such as (1) the impossibility of specifying a single, all-purpose measure of classification accuracy and (2) the selection of an appropriate sample size as well as the specification and the use of a measure of accuracy appropriate to the application.

A brief summary of OBIA-based object detection methods is given in Table 3.

Table 3 A brief summary of OBIA-based object detection methods

| Objects and publications | Strengths | Limitations |
|---|---|---|
| change detection (Bontemps et al., 2008; Chen and Hay, 2012; Contreras et al., 2016; Dissanska et al., 2009; Doxani et al., 2012; Doxani et al., 2008; Hussain et al., 2013; Im et al., 2008; Nebiker et al., 2014; Walter, 2004); land cover and land use mapping including vegetation, tree, water, residential, etc. (Baker et al., 2013; Benz et al., 2004; Blaschke, 2003; Blaschke et al., 2011; Blaschke et al., 2008; Contreras et al., 2015; D'Oleire-Oltmanns et al., 2014; De Pinho et al., 2012; Doleire-Oltmanns et al., 2013; Drăguţ and Blaschke, 2006; Drăguţ and Eisank, 2012; Duro et al., 2012; Eisank et al., 2011; Goodin et al., 2015; Hay et al., 2003; Hofmann et al., 2011; Kim et al., 2011; Leon and Woodroffe, 2011; Li et al., 2014; Li and Shao, 2013; Lisita et al., 2013; Macfaden et al., 2012; Mallinis et al., 2008; Mishra and Crews, 2014; Moskal et al., 2011; Myint et al., 2011; Phinn et al., 2012; Tzotsos et al., 2011; Walker and Blaschke, 2008; Walker and Briggs, 2007; Weng, 2009, 2011; Xie et al., 2008; Yu et al., 2006; Zhou, 2013; Zhou et al., 2009; Zhou and Troy, 2008); landslide mapping (Feizizadeh et al., 2014; Li et al., 2015b; Martha et al., 2010; Martha et al., 2011; Martha et al., 2012; Stumpf and Kerle, 2011). | The flexible incorporation of shape, texture, geometry, and contextual semantic features, as well as GIS-like functionality and expert knowledge, makes OBIA context-aware and multi-source capable. | Generic solutions to the full automation of segmentation process are still missing; The expert knowledge for how to define the classification rules are still subjective. |

## 6. Machine learning-based object detection

With the advance of machine learning techniques, especially the powerful feature representations and classifiers, many recent approaches regarded object detection as a classification problem and have achieved significant improvements. Fig. 5 gives the flowchart of machine learning-based object detection, in which object detection can be performed by learning a classifier that captures the variation in object appearances and views from a set of training data in a supervised or semi-supervised or weakly supervised framework. The input of the classifier is a set of regions (sliding windows or object proposals) with their corresponding feature representations and the output is their corresponding predicted labels, i.e., object or not. As can been seen from Fig. 5, feature extraction, feature fusion and dimension reduction (optional), and classifier training play the most important roles in the performance of object detection and hence we mainly focus on reviewing these three crucial steps.

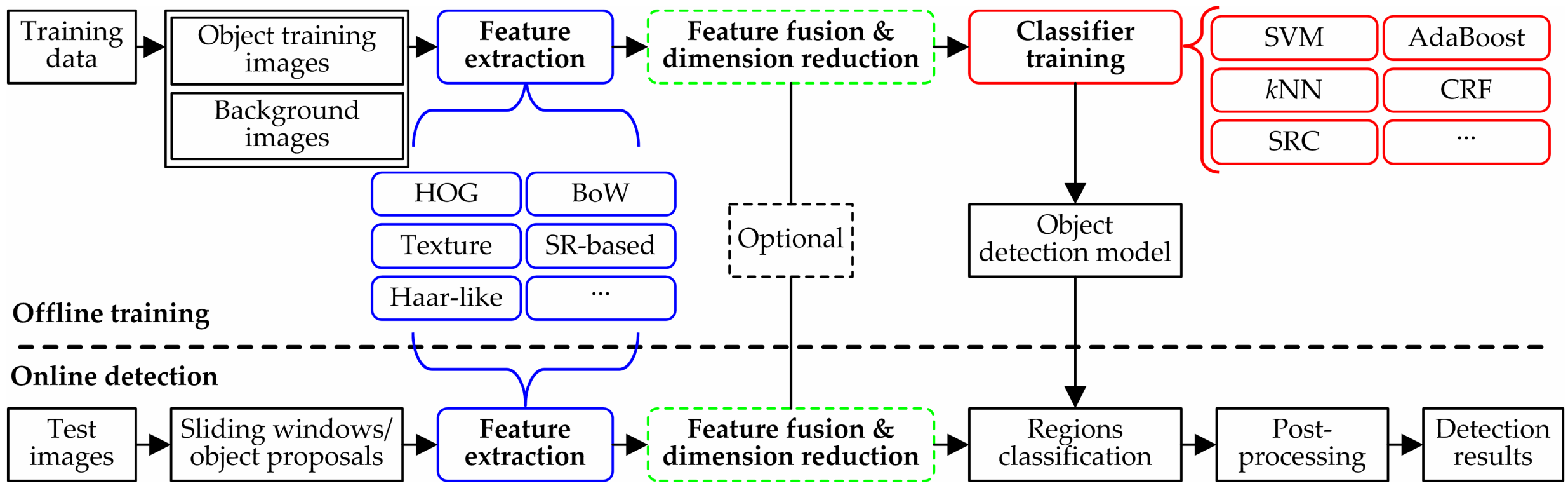

Fig. 5 The flowchart of machine learning-based object detection.

### 6.1 Feature extraction

Feature extraction is the mapping from raw image pixels to a discriminative high-dimensional data space. Since high-performance object detection is usually carried out in feature space, the feature extraction is quite important to construct high-performance object detection systems. Here we mainly focus on reviewing some widely used features for object detection in optical RSIs. Some reviews on feature extraction and feature coding in computer vision are available in (Andreopoulos and Tsotsos, 2013; Huang et al., 2014; Li et al., 2015c).

#### *6.1.1 Histogram of oriented gradients (HOG) feature and its extensions*

HOG feature was first proposed by Dalal and Triggs (2005) to represent objects by the distribution of gradient intensities and orientations in spatially distributed regions, which has been widely acknowledged as one of the best features to capture the edge or local shape information of the objects. Since its introduction, it has shown great success in many geospatial object detection algorithms (Cheng et al., 2013b; Cheng et al., 2014a, b; Grabner et al., 2008; Kembhavi et al., 2011; Tuermer et al., 2013). Besides, some part model-based methods (Cheng et al., 2013b; Cheng et al., 2014a, b; Zhang et al., 2014b; Zhang et al., 2015b) and the sparelets work (Cheng et al., 2015a; Cheng et al., 2015b; Cheng et al., 2015c) that built on the HOG feature have also shown impressive performance.

The implementation of HOG feature extraction can be briefly summarized as follows. 1) Calculate gradients at orthogonal directions with centered derivative mask $[-1,0,+1]$. 2) Compute gradient magnitudes and orientations for each pixel point. 3) Divide each image into fine and non-overlapping spatial regions called cells. 4) Accumulate a local one-dimensional histogram of gradient orientations over the pixels within each cell. The gradient at each pixel is quantized into one of nine orientation bins. Each pixel votes into the corresponding orientation bin with a voting weight based on the gradient magnitude. 5) Spatially connected cells are grouped into coarser blocks. 6) Run a normalization process on each block, by the division of the gradient energy inside the block, to provide strong illumination invariance and to reduce the sensitivity to gradient strength. Suppose $\boldsymbol{v}$ is the descriptor vector from a block, and the $\ell_p$ norm of $\boldsymbol{v}$ is used for normalization. Typical choices are

$$\ell_2\text{-norm: } \boldsymbol{v} \leftarrow \boldsymbol{v} / \sqrt{\|\boldsymbol{v}\|_2^2 + \xi^2}, \qquad \ell_1\text{-norm: } \boldsymbol{v} \leftarrow \boldsymbol{v} / \left(\|\boldsymbol{v}\|_1 + \xi\right) \tag{1}$$

where $\xi$ is a small constant and the results are insensitive to the value of $\xi$ over a large range. The HOG feature is the concatenation of normalized vectors in distributed blocks. Since histograms are computed over regions, they are fairly robust to some variability in the location of the parts of the object. The HOG descriptor is also invariant to rotations smaller than the size of the histogram orientation bin.

To further enhance the description capability of HOG for optical RSIs, several extensions are developed (Shi et al., 2014; Zhang et al., 2014b; Zhang et al., 2015b). In (Zhang et al., 2014b), rotation invariant HOG feature was achieved by rotating the coordinates of the descriptor and the gradient orientations according to the dominant orientation as SIFT descriptor (Lowe, 2004). Shi et al. (2014) developed a circle frequency-HOG feature for ship detection by combining circle frequency feature and HOG feature. Zhang et al. (2015b) introduced a strategy to generate a variant of HOG, namely HOG normalized by polar angle, to steer the rotation problem, in which the gradient direction of each pixel is normalized by its polar angle and then the HOG is generated according to the new directions.

#### *6.1.2 Bag-of-words (BoW) feature*

Among various features developed for visual object recognition, the BoW model (Li and Perona, 2005) is probably one of the most popular during the last decade. The main advantages of the BoW model is its simplicity, efficiency and invariance under viewpoint changes and background clutter, which was widely adopted by the community and results in good performance for geographic image classification (Xu et al., 2010; Yang and Newsam, 2010, 2011, 2013) and geospatial object detection (Bai et al., 2014; Cheng et al., 2013a; Sun et al., 2012; Zhang et al., 2015a).

There are generally five steps to construct BoW model and we will briefly state these steps. The first step is to detect keypoints from images. Many algorithms have been proposed for keypoint detection such as Harris-Laplacian detector (Mikolajczyk and Schmid, 2001) and difference of Gaussian (DoG) detector (Lowe, 2004). The second step is to compute local descriptors for the detected keypoints. There is a rich literature on local descriptors, among which SIFT descriptor (Lowe, 2004) is the most popular. The third step is to construct the visual vocabulary using clustering technique such as *k*-means over the computed descriptors. Given the dictionary, the fourth step is to vector quantize each keypoint into a visual word in the dictionary, a step that is often referred to as the encoding module. The last step of the BoW model is the pooling step that pools encoded local descriptors into a global histogram representation. Various pooling strategies have been proposed for the BoW model such as mean and max-pooling. Studies in (Lazebnik et al., 2006) have shown that it is important to take into account the spatial layout of keypoints in the pooling step. A well known example of this approach is spatial pyramid matching (SPM) (Lazebnik et al., 2006) which partitions the image into increasingly finer spatial sub-regions and constructs a histogram for each sub-region separately.

#### *6.1.3 Texture features*

Texture features aim to describe the local density variability and patterns inside the surface of an object, which are very important for identifying textural objects such as airport (Aytekin et al., 2013; Tao et al., 2011), buildings (Senaras et al., 2013), urban area (Zhong and Wang, 2007), vehicles (Eikvil et al., 2009; Grabner et al., 2008), and so on. Here we briefly review two commonly used texture features, namely Gabor feature and local binary patterns (LBP) feature.

**Gabor feature**. The Gabor feature (Jain et al., 1997) is a classical kind of texture features. It is computed by applying a bank of scale and orientation selective Gabor filters to an image. Gabor filters often appear in pairs as even and odd components with the following mathematical form

$$G_{\text{even}}(x,y)=\cos(k_x x+k_y y)\exp\left(-\frac{x^2+y^2}{2\sigma^2}\right),\ G_{\text{odd}}(x,y)=\sin(k_x x+k_y y)\exp\left(-\frac{x^2+y^2}{2\sigma^2}\right) \tag{2}$$

where $k_x$ and $k_y$ are spatial frequencies at that the filters respond most strongly. $\arctan\left(k_x/k_y\right)$ determines the orientation and $\sigma$ denotes the scale of the filters. A bank of Gabor filters is generated by selecting different combinations of $k_x$, $k_y$, and $\sigma$. Gabor filtering responds higher when the local texture changes at the same orientation with the same frequency. The Gabor feature dimension is determined by the image size and the filter bank size.

**LBP feature**. The LBP feature (Ojala et al., 2002) is a theoretically very simple yet efficient approach for texture description by computing the frequencies of local patterns in sub-regions. The local pattern is defined with a center pixel $p_c$, pixel number $N$ and radius $r$. A set of interpolated surrounding pixels $P=\{p_1,\cdots,p_N\}$ is generated with predefined $N$ and $r$. The values of surrounding pixels $P$ are then compared with the value of center pixel $p_c$ and a $N$-bit vector is generated. The LBP code is defined by the decimal value of the $N$-bit vector

$$LBP_{N,r}(p_c)=\sum_{i=1}^{N}2^{i-1}\phi\left(I(p_c),I(p_i)\right) \tag{3}$$

where $I(p_c)$ and $I(p_i)$ are the values of pixels $p_c$ and $p_i$, respectively. $\phi\left(I(p_c),I(p_i)\right)=1$ only if $I(p_c)<I(p_i)$, otherwise it is 0. By computing the decimal LBP code for each point, the frequencies of LBP codes appearing in spatially distributed cells can be recorded as the bin value of the LBP histogram. The bin number is $2^N$. Besides, by circularly right shifting the bitwise vector and calculating the minimal value for the histogram binning we can obtain rotation-invariant LBP.

#### *6.1.4 Sparse representation (SR)-based features*

Lately, with the development of compressed sensing theory, SR-based features have been extensively applied to hyperspectral image denoising (Zhao and Yang, 2015), hyperspectral image classification (Chen et al., 2011a, 2013; Qian et al., 2013; Tao et al., 2012), and object detection in RSIs (Chen et al., 2011b, c; Cheng et al., 2014c; Du and Zhang, 2014; Han et al., 2014; Liu and Shi, 2014; Yokoya and Iwasaki, 2015; Zhang et al., 2014a; Zhang et al., 2015c; Zhang et al., 2015d). The core idea of SR is to sparsely encode high-dimensional original signals by a few structural primitives in a low-dimensional manifold. The procedure of seeking the sparsest representation for test sample in terms of an over-complete dictionary endows itself with a discriminative nature to perform classification. The SR-based feature can be generally calculated via resolving a least-square-based optimization problem with constraints on the sparse coefficients

$$\boldsymbol{\alpha}=\arg\min_{\boldsymbol{\alpha}}\left\{\left\|\boldsymbol{x}-\mathbf{D}\boldsymbol{\alpha}\right\|_2^2+\lambda\varphi(\boldsymbol{\alpha})\right\} \tag{4}$$

where $\|\cdot\|_2$ denotes the $\ell_2$ norm, $\lambda$ is a scalar constant, $\boldsymbol{x}\in R^d$ is an input signal, $\mathbf{D}=[\boldsymbol{d}_1,\cdots,\boldsymbol{d}_M]\in R^{d\times M}$ is an over-complete dictionary with each column $\boldsymbol{d}_i\left(i=1,\cdots,M\right)$ representing an atom, and $\boldsymbol{\alpha}\in R^M$ is a sparse coefficient vector used as the feature of input signal $\boldsymbol{x}$. In general, the number of atoms $M$ is large, whereas the signal dimension $d$ is relatively small. The least-square term $\left\|\boldsymbol{x}-\mathbf{D}\boldsymbol{\alpha}\right\|_2^2$ pursues accurate reconstruction, i.e., a signal can be described with a small error, and the constraint term $\varphi(\boldsymbol{\alpha})$ pursues discriminative description, i.e., similar/different features obtain similar/different representations. The main difference among various SR-based features lies in the constraint term. A widely used constraint function is $\ell_1$ norm constraint $\varphi(\boldsymbol{\alpha})=\sum_{i=1}^{M}|\boldsymbol{\alpha}(i)|$. With the $\ell_1$ norm constraint, SR achieves the effect that similar signals share a part of dictionary.

In addition to $\ell_1$ norm constraint, there are many other SR-based feature extraction methods in the recent literature, such as discriminative sparse coding (Han et al., 2014), joint sparse coding (Chen et al., 2011b), SR-based Hough voting (Yokoya and Iwasaki, 2015), nonlinear SR (Zhang et al., 2015d), sparse transfer manifold embedding (Zhang et al., 2014a), Laplacian SR (Chen et al., 2011c), and class-specific and discriminative SR (Du and Zhang, 2014). All of them extend SR by substituting or modifying the constraint term. Due to limited space, we do not introduce them one by one here.

#### *6.1.5 Haar-like features*

Haar-like features were first introduced by Viola and Jones (2001) for face detection and now are widely used for object detection in RSIs such as (Grabner et al., 2008; Leitloff et al., 2010). Haar-like features can be defined by a set of rectangular regions which are either positive or negative weighted. The feature value is calculated as the sum of pixel values in rectangular regions with their corresponding weights. Haar-like feature can be mathematically formed as $\{(Rec_i, w_i) \mid i=1,\cdots,N\}$, where $N$ is the number of the rectangles, $Rec_i$ denotes the $i$-th rectangular region, and $w_i$ denotes the assigned weight. The feature value is computed as $Haar(Rec_i, w_i)=\sum_{i=1}^{N} w_i \cdot Sum(Rec_i)$, where $Sum(Rec_i)$ refers to the summed pixel values inside rectangle $Rec_i$. Besides, a normalization is required to reduce the effects of illumination changes. Suppose $\mu$ and $\sigma$ are the mean value and standard derivation, respectively, of pixel values inside local regions, then the normalized feature value is denoted as $(Haar(Rec_i, w_i)-\mu)/\sigma$.

### 6.2 Feature fusion and dimension reduction

After feature extraction, some optional techniques can be applied to further improve feature representation performance. We now briefly discuss two kinds of commonly used techniques, namely feature fusion and dimension reduction.

#### *6.2.1 Feature fusion*

Although many visual descriptors have been proposed to describe geospatial objects, as reviewed in Section 6.1, different features generally describe different aspects of the visual characteristics. Due to the diversity of object appearances, the fusion of heterogeneous visual features is somewhat necessary for high-performance object detection. Feature fusion motivates the integration of the information of multiple visual cues in order to generate more comprehensive feature representation. The fusion of complementary features can lead to significant improvement in object detection.

A simple and widely used feature fusion technique is known as linear vector concatenation. Briefly, suppose we have $N$ different kinds of feature vectors $\{\boldsymbol{v}_1, \boldsymbol{v}_2, \cdots, \boldsymbol{v}_N\}$, where each feature vector $\boldsymbol{v}_i$ represents a specific visual property, vector concatenation-based feature fusion is to connect all feature vectors into a new long-length feature vector $\boldsymbol{v}=[\boldsymbol{v}_1;\boldsymbol{v}_2;\cdots;\boldsymbol{v}_N]$. In some cases, normalization is required to avoid the bias caused by scale differences. For instance, in (Grabner et al., 2008), an on-line boosting algorithm is used to integrate and select features from Haar-like features, orientation histograms, and LBP features for comprehensive object representation. In (Zhong and Wang, 2007), the authors presented a multiple CRFs ensemble model to incorporate five types of texture features, namely gray-level occurrence feature, GLCM feature, Gabor feature, gradient orientation feature, and straight-line length feature, for urban area detection. Furthermore, some non-linear feature fusion methods were also proposed such as heterogeneous feature machines (HFM) (Cao et al., 2009) and sparse multimodal learning (SMML) approach (Wang et al., 2013).

#### *6.2.2 Dimension reduction*

Dimension reduction aims to generate compact low-dimensional features from high-dimensional ones, in order to reduce computational complexity. Compactness and distinctiveness are two important properties of dimension reduction. Dimension reduction techniques can be broadly classified into unsupervised and supervised methods. PCA is a classical linear unsupervised method, which aims to find and preserve projections with maximal variance for energy preservation. However, the generalization capability of unsupervised dimension reduction is limited because they are dataset-dependent. For the purpose of classification, supervised dimension reduction methods hold an advantage over unsupervised methods due to their use of the class information within the training data set, and hence have attracted more attention such as partial least squares (Kembhavi et al., 2011), Fisher discriminant analysis (Sugiyama, 2007), and linear discriminant analysis (LDA) (Hariharan et al., 2012). In (Sugiyama, 2007), interclass separation is emphasized by replacing the total covariance matrix in PCA by Fisher criterion. In (Hariharan et al., 2012), LDA is used to remove the correlations of HOG and obtain the Whitened Histograms of Orientation (WHO) features.

### 6.3 Classifier training

After feature extraction, feature fusion and dimension reduction, a classifier can be trained using a number of possible approaches with the objective of minimizing the misclassification error on the training dataset. In practice, many different learning approaches can be selected including, but are not limited to, support vector machine (SVM), AdaBoost, $k$-nearest-neighbor ($k$NN), conditional random field (CRF), sparse representation-based classification (SRC), and artificial neural network (ANN).

#### *6.3.1 Support vector machine (SVM)*

SVM is one of the most popular and effective machine learning algorithms for solving classification problems, which was proposed by (Vapnik and Vapnik, 1998) and now has been widely used for various object detection applications, such as man-made objects recognition (Inglada, 2007), road extraction (Das et al., 2011; Huang and Zhang, 2009; Song and Civco, 2004), change detection (Bovolo et al., 2008; De Morsier et al., 2013), multi-class object detection (Bai et al., 2014; Cheng et al., 2013b; Cheng et al., 2014a, b; Li and Itti, 2011), ship detection (Bi et al., 2012; Zhu et al., 2010), airport detection (Tao et al., 2011), airplane detection (Sun et al., 2012; Zhang et al., 2014b; Zhang et al., 2015b), weakly supervised object detection (Han et al., 2015; Zhang et al., 2015a). A recent review about the applications of SVMs in remote sensing can also be found in (Mountrakis et al., 2011).

In its simplest form, SVM is a linear binary classifier. For multiple classification problem, such as multi-class object detection tasks, adjustments are only made to the simple SVM binary classifier to operate as a multi-class classifier by using methods such as one-against-all and one-against-others. Furthermore, SVM can also be used as a non-linear classifier (called kernel SVM) to further improve the separation between classes by projecting the samples onto a higher dimensional feature space.

Taking a binary classification problem as an example, given a training set of $N$ samples $(X,Y)=\{(\boldsymbol{x}_1,y_1),\cdots,(\boldsymbol{x}_N,y_N)\mid \boldsymbol{x}_i\in R^d, i=1,\cdots,N\}$, where $\boldsymbol{x}_i$ is the feature vector corresponding to the $i$-th sample labeled by $y_i\in\{-1,+1\}$. For the linear and separable case, the classifier is trained by solving the following optimization problem

$$\min \tfrac{1}{2}\|\boldsymbol{w}\|^2 \quad \text{s.t.} \quad y_i\left(\langle \boldsymbol{x}_i,\boldsymbol{w}\rangle+b\right)\geq 1, \quad i=1,\cdots,N \tag{5}$$

where $\boldsymbol{w}\in R^d$ and $b\in R$ are to-be-learned classifier parameters, and $\langle \boldsymbol{x},\boldsymbol{y}\rangle$ denotes the inner product of $\boldsymbol{x}$ and $\boldsymbol{y}$. In order to deal with non-separable data, the soft-margin hyperplane is used. A set of slack variables $\xi_i$ is introduced to allow errors or data points inside the margin and a hyperparameter $C$ is used to tune the trade-off between the amount of accepted errors and the maximization of the margin. The optimization problem now becomes

$$\min \tfrac{1}{2}\|\boldsymbol{w}\|^2 + C\sum_{i=1}^{N}\xi_i \quad \text{s.t.} \quad y_i\left(\langle \boldsymbol{x}_i, \boldsymbol{w}\rangle + b\right) \geq 1-\xi_i,\ i=1,\cdots,N,\ \xi_i \geq 0 \tag{6}$$

Given a test sample $\boldsymbol{x}' \in R^d$, the SVM then classifies the test sample with the following decision function, which determines on which side of the separating hyperplane the sample $\boldsymbol{x}'$ lies.

$$f(\boldsymbol{x}') = \text{sign}\left(\langle \boldsymbol{x}', \boldsymbol{w}\rangle + b\right) \tag{7}$$

For non-linear classification problems, the data are first mapped into a higher dimensional feature space $S$ by $\boldsymbol{x} \in R^d \mapsto \boldsymbol{\Phi}(\boldsymbol{x}) \in S$ in which a separating hyperplane is built. This leads to the decision function becoming $f(\boldsymbol{x}') = \text{sign}\left(\langle \boldsymbol{\Phi}(\boldsymbol{x}'), \boldsymbol{w}\rangle + b\right)$, where $\boldsymbol{w}$ is now a new vector of $S$. However, non-linear SVM is sensitive to the selection of kernel. In the case when the appropriate kernel is selected with optimal parameters, SVM can give very good classification results.

### 6.3.2 *AdaBoost*

The AdaBoost algorithm (Freund, 1995; Freund and Schapire, 1996) is a widely used machine learning algorithm that combines many weak classifiers to form a strong classifier by adjusting the weights of training samples. In many object detection applications such as vehicle detection (Grabner et al., 2008; Leitloff et al., 2010; Tuermer et al., 2013), ship detection (Shi et al., 2014), and airport detection (Aytekin et al., 2013), the Adaboost classifier has shown good performance.

Various variants of AdaBoosting have been developed, such as Discrete AdaBoost (Freund and Schapire, 1997), Real AdaBoost (Schapire and Singer, 1999), and Gentle AdaBoost (Lienhart et al., 2003). It has been shown in (Lienhart et al., 2003) that Gentle AdaBoost outperforms both Discrete Adaboost and Real Adaboost due to its less sensitivity to errors in the training set. This property is very useful when working with RSIs since the acquisition of accurate annotation of training data in complicated scenes cannot always be guaranteed. Consequently, we here focus on a brief review of the Gentle AdaBoost algorithm. The basic algorithm works as follows: Given a training set $(X,Y) = \{(\boldsymbol{x}_1, y_1), \cdots, (\boldsymbol{x}_N, y_N) \mid \boldsymbol{x}_i \in R^d, y_i \in \{-1,+1\}, i=1,\cdots,N\}$, where $\boldsymbol{x}_i$ is the feature vector of $i$-th example and $y_i$ is its class label. The weights for all examples are initialized uniformly by $w_i = 1/N$. Based on the training set, a weak classifier $f_m(\boldsymbol{x})$ is trained by weighted least-squares fitting of the labels $Y$ to the feature vectors $X$ with weights $\boldsymbol{w}$. Then, the weights are updated by $w_i \leftarrow w_i \cdot \exp\left(-y_i f_m(\boldsymbol{x}_i)\right)/Z$ with the normalization factor $Z$ defined by $Z = \sum_{i=1}^{N} w_i$. At each boosting iteration a new weak classifier is added and the process is repeated until a certain stopping condition is met (e.g. a predefined number of $M$ weak classifiers are trained). Finally, a strong classifier $F(\boldsymbol{x})$ is computed as a linear combination of all weak classifiers

$$F(\boldsymbol{x}) = \sum_{m=1}^{M} f_m(\boldsymbol{x}) \tag{8}$$

In object detection, the sign of $F(\boldsymbol{x})$ decides the predicted object label, while its absolute value is a confidence measure for this prediction. The higher the absolute value is, the more confident is the result.

### 6.3.3 *k-nearest-neighbor (kNN)*

The *k*NN classifier is firstly proposed by Cover and Hart (1967). It is one of the most simple and traditional classification tools and has been extensively used for geospatial object detection and image classification (Blanzieri and Melgani, 2008; Cheng et al., 2013a; Haapanen et al., 2004; Ma et al., 2010; Yang et al., 2010; Zhu and Basir, 2005). It performs the classification by firstly consulting a reference set of labeled training samples.

Then, given a test sample, we identify a subset of $k$ training samples that are the closest to it. Finally, various decision strategies are adopted to classify the test sample and the most widely used strategy is assigning it to the class that appears most frequently within this subset.

Algorithmically, let $(X,Y)=\{(\boldsymbol{x}_1,y_1),\cdots,(\boldsymbol{x}_N,y_N)\,|\,\boldsymbol{x}_i\in R^d, y_i\in\{1,2,\cdots,C\}, i=1,\cdots,N\}$ be the training set with $N$ training samples, where $\boldsymbol{x}_i$ is the feature vector of the $i$-th sample and $y_i$ is its class label (assuming we have $C$ classes). Given a test sample $\boldsymbol{x}'\in R^d$ and an ordering function $\boldsymbol{f}_{\boldsymbol{x}'}$ (e.g. an Euclidean metrics $\boldsymbol{f}_{\boldsymbol{x}'}(\boldsymbol{x})=\|\boldsymbol{x}-\boldsymbol{x}'\|$), we can order the set of training samples $X$ to obtain its $k$ nearest neighbors $X'=\{\boldsymbol{x}'_1,\boldsymbol{x}'_2,\cdots,\boldsymbol{x}'_k\}$ with their corresponding labels denoted as $Y'=\{y'_1,y'_2,\cdots,y'_k\}$. The *k*NN then classifies the test sample $\boldsymbol{x}'$ with the following majority voting rule

$$j*=\arg\max_{j=1,\cdots,C}\sum_{i=1}^{k}\delta(y'_i,j) \tag{9}$$

where $\delta$ is the Kronecker delta defined as

$$\delta(i,j)=\begin{cases}0 & \text{if } i\neq j\\ 1 & \text{if } i=j\end{cases} \tag{10}$$

As can be seen from the above description, on the one hand, there are some attractive properties of the *k*NN classifier, such as the simplicity of the learning rule and the tuning of only one free parameter. On the other hand, the selection of the neighbor size $k$ is a challenging problem because different $k$ values will result in different performances. If $k$ is considerably huge, the neighbors searching will take large time whereas a small value of $k$ may decrease the prediction accuracy. To deal with this issue, different variants have been proposed in the literature (Blanzieri and Melgani, 2008; Dudani, 1976; Zhu and Basir, 2005).

#### *6.3.4 Conditional random field (CRF)*

CRF is introduced in (Lafferty et al., 2001) for labeling 1-D data. Kumar and Hebert (2003) further extended CRF to classify 2-D images. CRF has the advantage of incorporating the spatial contextual information to assign probabilities to the final label instead of only providing crisp decisions and hence has been applied successfully to various visual recognition tasks such as building detection (Li et al., 2015a; Wegner et al., 2011a; Wegner et al., 2011b), urban area detection (Zhong and Wang, 2007), and airport detection (Yao et al., 2015).

Here we briefly introduce CRF models concerning with binary classification task of distinguishing objects from background, such as the works in (Li et al., 2015a; Wegner et al., 2011a; Wegner et al., 2011b; Yao et al., 2015; Zhong and Wang, 2007). Given an input image, let $G=(S,E)$ be a graph denoting a CRF of $(\boldsymbol{x},\boldsymbol{y})$, where $\boldsymbol{x}=\{\boldsymbol{x}_i\}_{i\in S}$ is the observed data from the input image with $\boldsymbol{x}_i$ denoting the data from the $i$-th region, $\boldsymbol{y}=\{y_i\,|\,y_i\in\{-1,+1\}\}_{i\in S}$ is the corresponding labels at the image regions, $S$ is a set of nodes and each node corresponds to an image region data $\boldsymbol{x}_i$ and $E$ is a set of edges connecting certain pairs of neighboring nodes. In CRF framework, the random variables $\boldsymbol{x}$ and $\boldsymbol{y}$ are jointly distributed in a discriminative manner and the posterior distribution $P(\boldsymbol{y}\,|\,\boldsymbol{x})$ of the labels $\boldsymbol{y}$ given the observations $\boldsymbol{x}$ can be written as

$$P(\boldsymbol{y}\,|\,\boldsymbol{x})=\frac{1}{Z}\exp\left(\sum_{i\in S}A_i(\boldsymbol{x},y_i)+\sum_{i\in S}\sum_{j\in N_i}I_{ij}(\boldsymbol{x},y_i,y_j)\right) \tag{11}$$

where $Z$ is a normalizing constant known as the partition function, $A_i(\boldsymbol{x},y_i)$ is the association potential, $I_{ij}(\boldsymbol{x},y_i,y_j)$ is the interaction potential, and $N_i$ is the neighbors of the node $i$ in $G$.

The association potential $A_i(\boldsymbol{x},y_i)$ measures how likely a node $i$ is labeled $y_i$ with given the data $\boldsymbol{x}$, as in (Wegner et al., 2011a; Wegner et al., 2011b) which is generally defined as

$$A_i(\boldsymbol{x},y_i)=\exp\left(y_i\boldsymbol{w}^{\mathrm{T}}\boldsymbol{h}_i(\boldsymbol{x})\right) \tag{12}$$

where $\boldsymbol{h}_i(\boldsymbol{x})$ denotes the feature representation vector of node $i$ and $\boldsymbol{w}^{\mathrm{T}}$ is a weight vector containing the weights of the features in $\boldsymbol{h}_i(\boldsymbol{x})$ that are learned during the training process.

The interaction potential $I_{ij}(\boldsymbol{x}, y_i, y_j)$ describes how two nodes $i$ and $j$ interact, as in (Wegner et al., 2011a; Wegner et al., 2011b) which is defined as

$$I_{ij}(\boldsymbol{x}, y_i, y_j) = \exp\left(y_i y_j \boldsymbol{v}^{\mathrm{T}} \boldsymbol{\mu}_{ij}(\boldsymbol{x})\right) \tag{13}$$

where $\boldsymbol{\mu}_{ij}(\boldsymbol{x})$ denotes a feature vector which is simply calculated by either subtracting or concatenating $\boldsymbol{h}_i(\boldsymbol{x})$ and $\boldsymbol{h}_j(\boldsymbol{x})$, and vector $\boldsymbol{v}^{\mathrm{T}}$ contains the weights of the features $\boldsymbol{\mu}_{ij}(\boldsymbol{x})$ that are adjusted during the training process. It should be noted that the interaction potential $I_{ij}(\boldsymbol{x}, y_i, y_j)$ is not only a function of adjacent labels $y_i$ and $y_j$ in the local neighborhood, but of all data, too. This means that the interaction potential can effectively incorporate both local and global context information which is actually a major advantage concerning automatic analysis of remote sensing data.

#### *6.3.5 Sparse representation-based classification (SRC)*

SRC was first proposed by Wright to tackle face recognition problem (Wright et al., 2009), which aims at looking for the sparsest representation of a test example in a dictionary composed of all training data across all classes. In the last few years, SRC and its variants have been widely used in RSI analysis field, including geospatial object detection (Chen et al., 2011b, c; Cheng et al., 2014c; Han et al., 2014; Liu and Shi, 2014; Zhang et al., 2015c; Zhang et al., 2015d) and hyperspectral image classification (Chen et al., 2011a; Qian et al., 2013; Tao et al., 2012), and have shown excellent performances.

A brief review of SRC is as follows. Assume we have $C$ classes and we are given $m_i$ training data $\left\{\boldsymbol{a}_{ij} \in R^d\right\}_{j=1}^{m_i}$ for each class $i (i = 1, \cdots, C)$. We denote by $\mathbf{A}_i = \left[\boldsymbol{a}_{i1}, \cdots, \boldsymbol{a}_{im_i}\right] \in R^{d \times m_i}$ the collection of training data of the $i$-th class and denote by $\mathbf{A} = \left[\mathbf{A}_1, \cdots, \mathbf{A}_C\right]$ the collection of all training data across all classes. Given a test example $\boldsymbol{x} \in R^d$, which belongs to one of the $C$ classes, our goal is to find the class to which the test example belongs. The SRC method is based on the idea that a test example has a sparse representation in the dictionary of all the training data across different classes. Therefore, in principle, we can sparsely code $\boldsymbol{x}$ on $\mathbf{A}$ via solving the following optimization problem

$$\boldsymbol{\alpha} = \arg\min_{\boldsymbol{\alpha}} \left\{ \left\| \boldsymbol{x} - \mathbf{A}\boldsymbol{\alpha} \right\|_2^2 + \lambda \left\| \boldsymbol{\alpha} \right\|_1 \right\} \tag{14}$$

We can then determine the class of a given test example as the class that best represents the test example using its training data. More precisely, for a given test example $\boldsymbol{x}$, we define the $i$-th residual, i.e., error between the test example and the reconstruction from training samples in the $i$-th class, to be

$$r_i(\boldsymbol{x}) = \left\| \boldsymbol{x} - \mathbf{A}_i \hat{\boldsymbol{\alpha}}_i \right\|_2, \quad i = 1, \cdots, C \tag{15}$$

where $\hat{\boldsymbol{\alpha}} = [\hat{\boldsymbol{\alpha}}_1; \cdots; \hat{\boldsymbol{\alpha}}_C]$ is the optimal solution of (14) and $\hat{\boldsymbol{\alpha}}_i$ is the coefficient vector associated with the $i$-th class. The class of $\boldsymbol{x}$ is determined as the one with the minimal residual

$$\mathrm{class}(\boldsymbol{x}) = \arg \min_{i=1,\cdots,C} r_i(\boldsymbol{x}) \tag{16}$$

#### *6.3.6 Artificial neural network (ANN)*

ANN, known as a kind of pattern classifiers, was proposed in the early 1980s. It has the capability to learn patterns whose complexity makes them difficult to analyze using other conventional approaches. Some typical neural network architectures, such as multilayer perceptron (MLP) (Li et al., 2001; Mokhtarzade and Zoej, 2007; Pacifici et al., 2009), hopfield neural networks (Ghosh et al., 2007), extreme learning machine (ELM) (Malek et al., 2014; Tang et al., 2015), and convolutional neural network (CNN) (Jin and Davis, 2007; Wang et al., 2015), have been successfully used in many remote sensing applications including ship detection (Tang et al., 2015), vehicle detection (Jin and Davis, 2007), road detection (Mokhtarzade and Zoej, 2007; Wang et al., 2015), tree detection (Malek et al., 2014), fire smoke detection (Li et al., 2001), change detection (Ghosh et al., 2007), and land-use classification (Pacifici et al., 2009). Here, we briefly review the structure of neural networks. For more details and the latest advances, readers can refer to (Bishop, 1995; LeCun et al., 2015).

Neural networks are made up of a large number of simple processing units called nodes or neurons. The main task of a neuron is to receive input from its neighbors, to compute an output and to send the output to its neighbors. Neurons are usually organized into layers with full or random connections between successive layers. An ANN generally consists of three types of layers, namely input layer, hidden layer, and output layer, that receive, process and present the final results, respectively. There are two main stages in the operation of an ANN classifier, i.e., learning (training) and recalling. Learning is the process of adapting or modifying the connection weights so that the network can fulfill a specific task. This process is mainly performed with a supervised learning algorithm using a training set, in which random weights are first given at the beginning of training, and then the algorithm performs weights tuning by minimizing the error of misclassification. Recalling asks how the network can operate based on what it has learned in the training stage. It actually uses the trained network for interpolation and extrapolation, such as classification and regression.

#### *6.3.7 Other classifiers*

Besides the aforementioned classifiers, some other machine learning algorithms for object detection in optical RSIs are also used, such as Gaussian mixture model (GMM) (Ari and Aksoy, 2014; Bhagavathy and Manjunath, 2006), Markov random fields (MRF) (Benedek et al., 2015; Kasetkasem and Varshney, 2002; Le Hégarat-Mascle and André, 2009), random forest (Dong et al., 2015; Yin et al., 2015), texton forests (Lei et al., 2012, 2014), support tensor machine (STM) (Zhang et al., 2011b), Bayesian minimum risk classifier (Zhao and Nevatia, 2003), partial least squares classifier (Kembhavi et al., 2011), logistic regression classifier (Corbane et al., 2010; Zhang and Couloigner, 2006), LDA (Eikvil et al., 2009), quadratic discriminant analysis (QDA) (Eikvil et al., 2009), and decision fusion (Fauvel et al., 2006; Senaras et al., 2013). Readers can refer to the relevant publications for details.

A brief summary of machine learning-based object detection methods is given in Table 4.

Table 4 A brief summary of machine learning-based object detection methods

| Objects and publications | Strengths | Limitations |
|---|---|---|
| airplane (Lei et al., 2012; Liu and Shi, 2014; Sun et al., 2012; Zhang et al., 2014b; Zhang et al., 2015b); airport (Aytekin et al., 2013; Cheng et al., 2014c; Tao et al., 2011; Yao et al., 2015); building (Ari and Aksoy, 2014; Lei et al., 2012; Li et al., 2015a; Senaras et al., 2013; Wegner et al., 2011a; Wegner et al., 2011b); change detection (Benedek et al., 2015; Bovolo et al., 2008; De Morsier et al., 2013; Ghosh et al., 2007; Kasetkasem and Varshney, 2002; Lei et al., 2014); forest (Haapanen et al., 2004); landslide (Cheng et al., 2013a); multi-class objects (e.g. airport, airplane, ship, storage tank, etc.) (Bai et al., 2014; Cheng et al., 2013b; Cheng et al., 2014a, b; Han et al., 2015; Han et al., 2014; Li and Itti, 2011; Yokoya and Iwasaki, 2015; Zhang et al., 2015a; Zhou et al., 2015a); road (Das et al., 2011; Huang and Zhang, 2009; Inglada, 2007; Mokhtarzade and Zoej, 2007; Song and Civco, 2004; Wang et al., 2015; Zhang and Couloigner, 2006); ship (Bi et al., 2012; Corbane et al., 2010; Shi et al., 2014; Tang et al., 2015; Zhu et al., 2010); tree (Malek et al., 2014); land-use/land-cover (Bhagavathy and Manjunath, 2006; Fauvel et al., 2006; Pacifici et al., 2009; Zhong and Wang, 2007); vehicle (Eikvil et al., 2009; Grabner et al., 2008; Jin and Davis, 2007; Kembhavi et al., 2011; Leitloff et al., 2010; Tuermer et al., 2013; Zhao and Nevatia, 2003). | Object model (detector) can be automatically established via machine learning technique; Detection system is scalable and compatible; Have high detection accuracy. | Need a lot of training samples of objects and non-objects to learn classifiers; Detection accuracy depends on the training samples. |

## 7. Datasets and evaluation metrics

During the last decades, many efforts have been made to develop various methods for object detection from aerial and satellite images. In order to make the results of such algorithms more comparable, it is of paramount importance to introduce some publicly available benchmarking datasets and some standard evaluation metrics to help researchers conduct further studies and compare different algorithms. Here we summarize five publicly available datasets and three standard evaluation metrics.

### 7.1 Datasets

**NWPU VHR-10 dataset** (Cheng et al., 2014a)[1]. This is a challenging 10-class geospatial object detection dataset, which can be used for both single class and multi-class objects detection. These ten classes of objects are airplane, ship, storage tank, baseball diamond, tennis court, basketball court, ground track field, harbor, bridge, and vehicle. This dataset contains totally 800 VHR optical RSIs, from which 757 airplanes, 302 ships, 655 storage tanks, 390 baseball diamonds, 524 tennis courts, 159 basketball courts, 163 ground track fields, 224 harbors, 124 bridges, and 477 vehicles were manually annotated with axis-aligned bounding boxes used for ground truth. 715 color images were acquired from Google Earth with the spatial resolution ranging from 0.5-m to 2-m, and 85 pansharpened color infrared (CIR) images were acquired from Vaihingen data with a spatial resolution of 0.08-m. The Vaihingen data was provided by the German Society for Photogrammetry, Remote Sensing and Geoinformation (DGPF) (Cramer, 2010): http://www.ifp.uni-stuttgart.de/dgpf/DKEPAllg.html.

**SZTAKI-INRIA building detection dataset** (Benedek et al., 2012)[2]. This dataset is an excellent resource for benchmarking building extraction algorithms. It contains the rectangular footprints of 665 buildings in nine aerial or satellite images taken from Budapest and Szada (both in Hungary), Manchester (U.K.), Bodensee (Germany), and Normandy and Cot d'Azur (both in France), as well as manually annotated ground truth data. Two images (Budapest and Szada) are aerial images, and the remaining images are satellite images acquired from Google Earth, IKONOS, and QuickBird. All of the images contain only RGB information.

**TAS aerial car detection dataset** (Heitz and Koller, 2008)[3]. This dataset consists of 30 images extracted from Google Earth and a total of 1319 manually labeled cars with axis-aligned bounding boxes used for ground truth. The images are color and of size 792×636. The average car window is approximately 45×45 pixels.

**Overhead imagery research dataset (OIRDS)** (Tanner et al., 2009)[4]. The OIRDS is designed for vehicle detection algorithms. It is a large collection of about 900 overhead images, captured using aircraft-mounted cameras. The total number of vehicles annotated in the data set is around 1,800. Ground sample distance (GSD) values for the OIRDS images range from a low of 0.0838-m to a high of 0.3048-m.

**IITM road extraction dataset** (Das et al., 2011)[5]. This dataset consists of 200 multispectral satellite images of size 512×512 each, where 100 images are of developed countries and 100 images are of emerging countries. They were created from satellite images with 1-m spatial resolution from Wikimapia (http://www.wikimapia.org/). With the advice (used visual observation and geolocation) of a GIS expert, this dataset was divided into four categories with 50 images per category. For each image in the dataset, a ground truth road map was also obtained using a human operator.

[1] http://pan.baidu.com/s/1hqwzXeG
[2] http://web.eee.sztaki.hu/remotesensing/building_benchmark.html
[3] http://ai.stanford.edu/~gaheitz/Research/TAS/tas.v0.tgz
[4] http://sourceforge.net/apps/mediawiki/oirds
[5] http://www.cse.iitm.ac.in/~vplab/satellite.html

### 7.2 Evaluation metrics

Here, we explain three universally-agreed, standard, and easy-to-understand measures for evaluating the object detection methods, namely precision-recall curve, F-measure, and average precision. The first evaluation metric is based on the overlapping area between detections and ground truth. From this metric, we also report the F-Measure, which jointly considers recall and precision, and average precision, which is the area under the precision-recall curve.

**Precision-recall curve (PRC)**. The $Precision$ measures the fraction of detections that are true positives and the $Recall$ measures the fraction of positives that are correctly identified. Let $TP$, $FP$, and $FN$ denote the number of true positives, the number of false positives, and the number of false negatives. The $Precision$ and $Recall$ can be formulated as

$$Precision = TP/(TP+FP) \tag{17}$$

$$Recall = TP/(TP+FN) \tag{18}$$

The PRC measure can evaluate the performance of an object detection algorithm at both pixel-level and object-level. For the pixel-level evaluation, true positives refer to the pixels assigned as target objects in both ground truth and detection result, false positives refer to the pixels assigned as target objects in detection result but not in ground truth, and false negatives refer to the pixels assigned as a target object in ground truth but not in detection result. In an object-level evaluation, a detection is labeled as true positive if the area overlap ratio $a_o$ between the detection and ground truth object exceed a predefined threshold $\lambda$ by the formula

$$a_o = \frac{area(detection \cap ground_truth)}{area(detection \cup ground_truth)} > \lambda \tag{19}$$

where $detection \cap ground_truth$ denotes the intersection of the detection and ground truth object and $detection \cup ground_truth$ denotes their union. Otherwise the detection is considered as a false positive. In addition, if several detections overlap with a same ground truth object, only one is considered as true positive and the others are considered as false positives.

**F-measure**. Often, neither $Precision$ nor $Recall$ can comprehensively evaluate the quality of an object detection method. To this end, the F-measure is proposed to combine the $Precision$ and $Recall$ metrics to a single measure by weighted harmonic mean of them with a non-negative weight $\beta^2$

$$F_\beta = \frac{(1+\beta^2)Precision \times Recall}{\beta^2 Precision + Recall} \tag{20}$$

Setting $\beta^2 < 1$ can raise more importance to the $Precision$ value and vice versa. As suggested by many object detection works (Benedek et al., 2012; Han et al., 2015; Han et al., 2014; Li et al., 2015a; Ok, 2013; Ok et al., 2013), $\beta$ is often set to 1 to give each an equal importance, which results in $F_\beta = F_1$.

**Average precision (AP)**. The AP computes the average value of $Precision$ over the interval from $Recall = 0$ to $Recall = 1$, i.e., the area under the PRC, so the higher the AP value is, the better the performance and vice versa.

## 8. Promising research directions

Here we propose two promising research directions, namely deep learning-based feature representation and weakly supervised learning-based geospatial object detection, to further advance the development of object detection task.

### 8.1 Deep learning-based feature representation

Currently, most of geospatial object detection methods, except for few works such as (Han et al., 2015; Tang et al., 2015; Wang et al., 2015; Zhou et al., 2015b), are still dominated by handcrafted features and shallow learning-based features (e.g. SR-based features (Chen et al., 2011b, c; Cheng et al., 2014c; Du and Zhang, 2014; Han et al., 2014; Liu and Shi, 2014; Yokoya and Iwasaki, 2015; Zhang et al., 2014a; Zhang et al., 2015c; Zhang et al., 2015d)). Although these features have shown impressive success for some specific object detection tasks, the involvement of human ingenuity in feature design or shallow structure significantly influence the representational power as well as the effectiveness for object detection. Especially as the visual recognition task becomes more challenging, the description capability of those features may become limited or even impoverished.

In 2006, a breakthrough in feature learning and deep learning was made by Hinton and Salakhutdinov (2006). Since then, deep learning has shown its stronger feature representation power in a wide range of computer vision applications. On the one hand, compared with handcrafted features which involves in abundant human ingenuity for features design, deep learning features are directly extracted from data in an automated fashion via neural networks of deep architecture. Thus, the burden for human engineering-based feature design has been transferred to the network construction. On the other hand, compared with traditional shallow learning techniques for feature representations (e.g. sparse coding), the deep architecture of neural networks can extract much more powerful feature representation. Besides, the feature maps from higher layers of the deep neural network display semantic abstracting properties. The representation power of deep learning feature can be significantly strengthened with stacked layers of neural networks.

Despite the progress made so far, there are also two notable challenges for deep learning-based feature representation. First, the deep learning features are much dependent on big image data because the deep neural networks need to be trained on a big enough training data set with abundant and diverse images to avoid overfitting. How to reduce the dependency to large-scale training data is a challenging issue. Second, the time cost for feature extraction with deep neural networks for online object detection is still very large. Many efforts are required to further improve the computational efficiency.

### 8.2 Weakly supervised learning-based geospatial object detection

With the explosive growth of satellite and aerial images in both quantity and quality, supervised learning-based object detection methods encounter a serious challenge caused by the fact that these approaches often require a large number of training samples with accurate manual annotation to train effective detectors. However, since VHR images often contain a greater range of object categories, big to an airport, small to a vehicle, manual annotation of various objects in large image sets is hardly achievable due to its expensive, time consuming and ineffective nature.

A few efforts (Bai et al., 2014; Capobianco et al., 2009; Cheng et al., 2014a; De Morsier et al., 2013; Liu et al., 2008) have been made to alleviate the work of human annotation. One common idea is to adopt the semi-supervised learning methods. Such methods apply a self-learning or an active learning technique where machine learning algorithms can automatically explore the most informative unlabeled examples based on a limited set of available labeled examples. Then, these selected unlabeled examples are combined with the initial labeled examples for the training of object detector or classifier. Although semi-supervised learning methods can considerably reduce the labor of human annotation, they still inevitably require a number of manually labeled training examples.

Weakly supervised learning (WSL) is more desirable for further reducing the human annotation significantly, in which the training set needs only binary labels indicating whether an image contains the target object or not. Although a few WSL approaches have been developed for natural scene image analysis (Deselaers et al., 2012; Zhu and Shao, 2014), those methods cannot be directly used for geospatial object detection because they are insufficient to address the challenges of RSIs, such as variations in the visual appearance of objects and complex background clutter. As an initial effort, our recent work in (Han et al., 2015; Zhang et al., 2015a; Zhou et al., 2015a; Zhou et al., 2015b) have shown the feasibility of using WSL for geospatial object detection. However, two challenging problems still exist. The first problem is the existing WSL-based geospatial object detection methods only work for a specific object class. Generally, a large-scale RSI often contains multiple classes of interesting objects instead of only a single one. Therefore, it is of great significance to develop a WSL-based object detection framework that can simultaneously identify multiple classes of objects. The other problem is the performance of existing WSL-based methods is still far from satisfactory. It still needs many efforts to develop more effective methods to improve the detection accuracy.

## 9. Conclusion

Object detection in optical RSIs has always been a fundamental but challenging issue in the field of aerial and satellite image analysis. During the last decades, considerable efforts have been made to develop various methods for the detection of different types of objects. In this paper, a review of the recent progress in this field was presented. We broadly categorized the methods into three groups, namely template matching-based methods, knowledge-based methods, and machine learning-based methods, and exhaustively reviewed them respectively. Five publicly available datasets and three standard evaluation metrics were also summarized. Besides, we also discussed the challenges of current studies and proposed two promising research directions in future. This survey will be significantly beneficial for the researchers to better understand this research field.

## Acknowledgements

The authors would like to thank the associate editor and all anonymous reviewers for their constructive comments. This work was partially supported by the National Science Foundation of China under Grants 61401357 and 61473231, the China Postdoctoral Science Foundation under Grants 2014M552491 and 2015T81050, and the Aerospace Science Foundation of China under Grant 20140153003.